\documentclass{article}
\usepackage[utf8]{inputenc}
\usepackage[T1]{fontenc}
\usepackage{graphicx}
\usepackage{booktabs}
\usepackage{epstopdf}
\usepackage{hyperref}
\usepackage{geometry}
\usepackage{textcomp}
\usepackage[utf8]{inputenc}
\DeclareUnicodeCharacter{2212}{\ensuremath{-}}

\usepackage{graphicx}

\usepackage{tikz}
\usepackage{makecell}
\usepackage[dvipsnames]{xcolor}

\usepackage{authblk}

\usepackage[table]{xcolor}
\usepackage{colortbl}
\usepackage{pgf} 

\newcommand{\divcell}[3]{%
  \begingroup
  \pgfmathsetmacro{\mid}{(#2+#3)/2}%
  \pgfmathparse{#1>\mid?1:0}\edef\ispos{\pgfmathresult}%
  \pgfmathsetmacro{\amp}{100*abs((#1-\mid)/max(#3-\mid,1e-9))}%
  \pgfmathsetmacro{\amp}{max(min(\amp,100),0)}%
  \ifnum\ispos=1
    \cellcolor{red!\amp!white}+#1\%%
  \else
    \cellcolor{blue!\amp!white}#1\%%
  \fi
  \endgroup
}

\newcommand{\divcellSlope}[3]{%
  \begingroup
  \pgfmathsetmacro{\mid}{(#2+#3)/2}%
  \pgfmathparse{#1>\mid?1:0}\edef\ispos{\pgfmathresult}%
  \pgfmathsetmacro{\amp}{100*abs((#1-\mid)/max(#3-\mid,1e-9))}%
  \pgfmathsetmacro{\amp}{max(min(\amp,100),0)}%
  \ifnum\ispos=1
    \cellcolor{red!\amp!white}+#1
  \else
    \cellcolor{blue!\amp!white}#1
  \fi\ (pp/yr)%
  \endgroup
}

\usepackage{tabularx,booktabs,array}
\newcolumntype{L}[1]{>{\raggedright\arraybackslash}p{#1}}
\newcolumntype{Y}{>{\raggedright\arraybackslash}X}

\newcolumntype{L}[1]{>{\raggedright\arraybackslash}p{#1}}
\newcolumntype{Y}{>{\raggedright\arraybackslash}X}

\title{Vision Language Models: A Survey of 26K Papers (CVPR, ICLR, NeurIPS 2023–2025)}
\author[1]{Fengming Lin\thanks{Email: fengming.lin@manchester.ac.uk; sdulinfm@gmail.com}}
\affil[1]{School of Computer Science, The University of Manchester, Manchester, UK}

\begin{document}

\maketitle

\begin{abstract}
We present a transparent, reproducible measurement of research trends across 26,104 accepted papers from CVPR, ICLR, and NeurIPS spanning 2023–2025. Titles and abstracts are normalized, phrase-protected, and matched against a hand-crafted lexicon to assign up to 35 topical labels and mine fine-grained cues about tasks, architectures, training regimes, objectives, datasets, and co-mentioned modalities. The analysis quantifies three macro shifts: (1) a sharp rise of multimodal vision–language–LLM work, which increasingly reframes classic perception as instruction following and multi-step reasoning; (2) steady expansion of generative methods—with diffusion research consolidating around controllability, distillation, and speed; and (3) resilient 3D and video activity, with composition moving from NeRFs to Gaussian splatting and a growing emphasis on human- and agent-centric understanding. Within VLMs, parameter-efficient adaptation (prompting, adapters/LoRA) and lightweight vision–language bridges dominate; training practice shifts from building encoders from scratch to instruction tuning and finetuning strong backbones; contrastive objectives recede relative to cross-entropy/ranking and distillation. Cross-venue comparisons show CVPR’s stronger 3D footprint and ICLR’s highest VLM share, while reliability themes (efficiency, robustness) diffuse across areas. We release the lexicon and methodology to enable auditing and extension. Limitations include lexicon recall and abstract-only scope, but the longitudinal signals are consistent across venues and years.
\end{abstract}

\section{Introduction}
The computer‑vision and machine‑learning community has undergone a visible transition in 2023–2025. 
With the consolidation of large‑scale pretrained models (CLIP/BLIP/LLaVA \cite{radford2021clip, li2022blip, liu2023llava} families, ViT backbones) and ubiquitous diffusion‑based generators, the research focus has shifted: classic perception remains active, yet a large fraction of accepted papers are now organized around multimodal training, general‑purpose reasoning or generation, and efficiency. 
While many reports provide anecdotal evidence, our goal is to quantify this transition from primary sources: the official titles and abstracts of accepted papers at CVPR, ICLR, and NeurIPS.

\section{Data and Methodology}

\textbf{Data:} 
We ingest all JSONL files collected by our Python spider: CVPR (2023: 2,353 papers; 2024: 2,713; 2025: 2,871), ICLR (2023: 4,372; 2024: 2,260; 2025: 3,704), and NeurIPS (2023: 3,337; 2024: 4,494). After removing empty records, 26,104 abstracts remain for analysis. In addition, for trend analysis only, we also include 8,424 papers from 2022; these 2022 records are used solely to study longitudinal trends and are excluded from content analysis, which focuses on the most recent three years.
\newline
\textbf{Text processing:} 
We normalize Unicode, lowercase, strip punctuation, protect multi‑word phrases (e.g., ``gaussian splatting'', ``neural radiance fields'', ``vision language model'') as single tokens, and remove general stopwords and generic CV terms to retain the technical content. 
\newline
\textbf{Labeling:} 
Each abstract is matched against 35 regular‑expression categories (Diffusion, Vision‑Language/LLM, 3D, Video, Robustness, Efficiency, etc.). A paper can receive multiple labels. We then compute prevalence as the fraction of abstracts in a year that match the label. 
\newline
\textbf{Fine‑grained mining:} 
For selected areas we search for sub‑topics: tasks (e.g., grounding), architecture motifs (e.g., LoRA/adapters), training regimes (pretrain+fine‑tune, instruction tuning), representative losses, and named datasets. 
\newline
\textbf{Caveat:} 
The approach is transparent and replicable, but lexicon‑driven—precision is high for canonical phrases; recall may miss niche synonyms; all numbers refer to abstracts only.

\section{Macro Trends}
\begin{figure}[h]
  \centering
  \includegraphics[width=\linewidth]{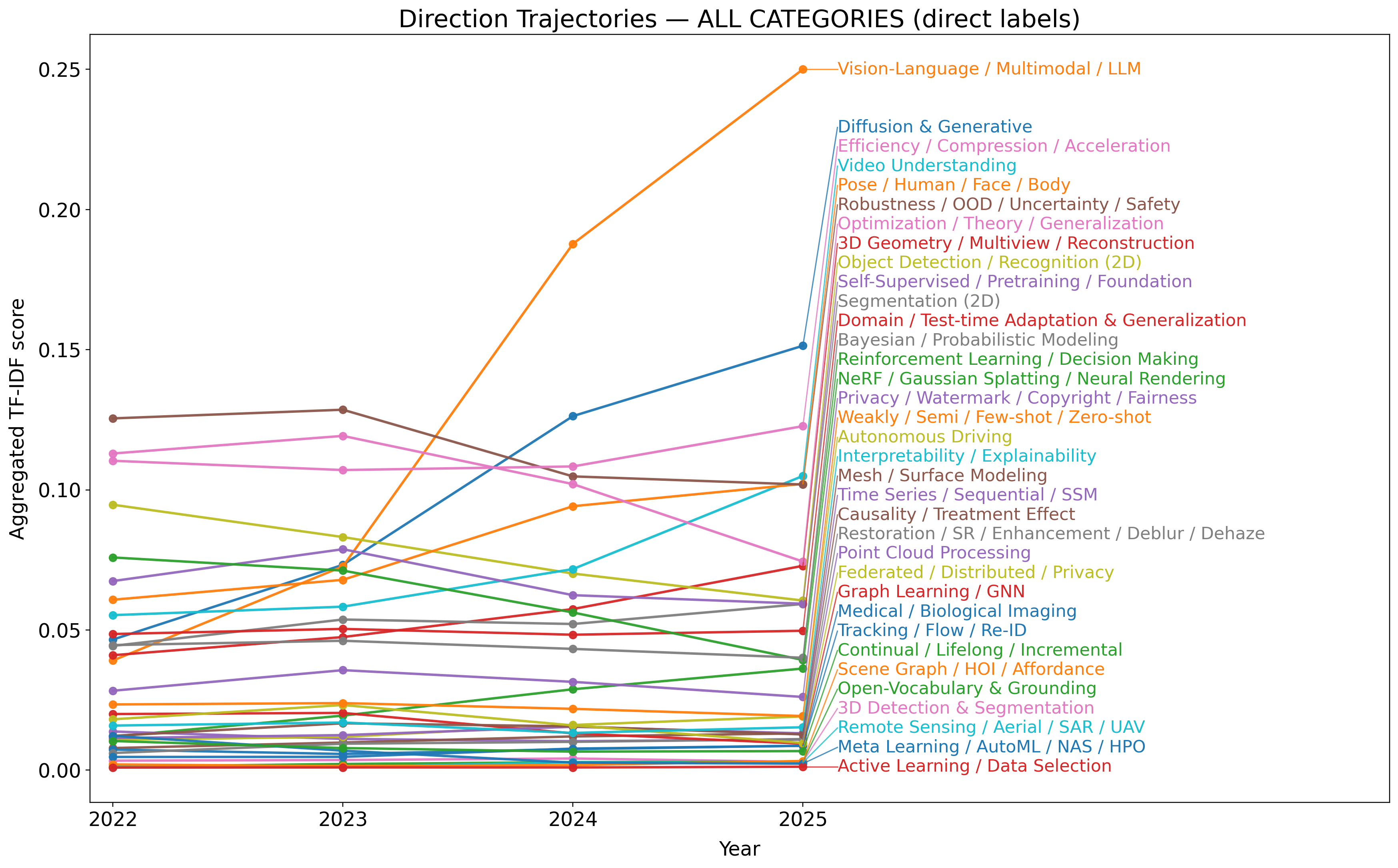}
  \caption{\textbf{Direction Trajectories across CVPR+ICLR+NeurIPS — ALL CATEGORIES (direct labels).}
  Each curve is the yearly aggregated TF--IDF mass for a direction (integer year ticks).}
  \label{fig:all-lines-all}
\end{figure}

Fig~\ref{fig:all-lines-all} summarizes the fraction of papers per year tagged as VLM/LLM, diffusion, 3D (NeRF/Gaussian + geometry), and video understanding et al. 
The rise of VLM is unmistakable: from 16\% (2023) to 40\% (2025) of all abstracts we analyzed. By 2025, the VLM share reaches 39.5\% at CVPR and 40.7\% at ICLR. Diffusion expands in parallel (8\%$\rightarrow$14.9\%$\rightarrow$19.2\%), while classic 3D remains stable overall with composition shifting from NeRFs to Gaussian splatting. Video has a steady incline, partly due to video‑LLMs and long‑context modeling.

The consolidated view highlights three macro patterns:
(i) a sharp takeoff of \emph{VLM/LLM}; 
(ii) sustained growth of \emph{Generative Model}, with increasing integration into perception pipelines; and 
(iii) steady increases in \emph{Video Understanding} and stable but reconfigured \emph{3D Reconstruction} trajectories.

\begin{figure}[h]
  \centering
  \includegraphics[width=16cm]{\detokenize{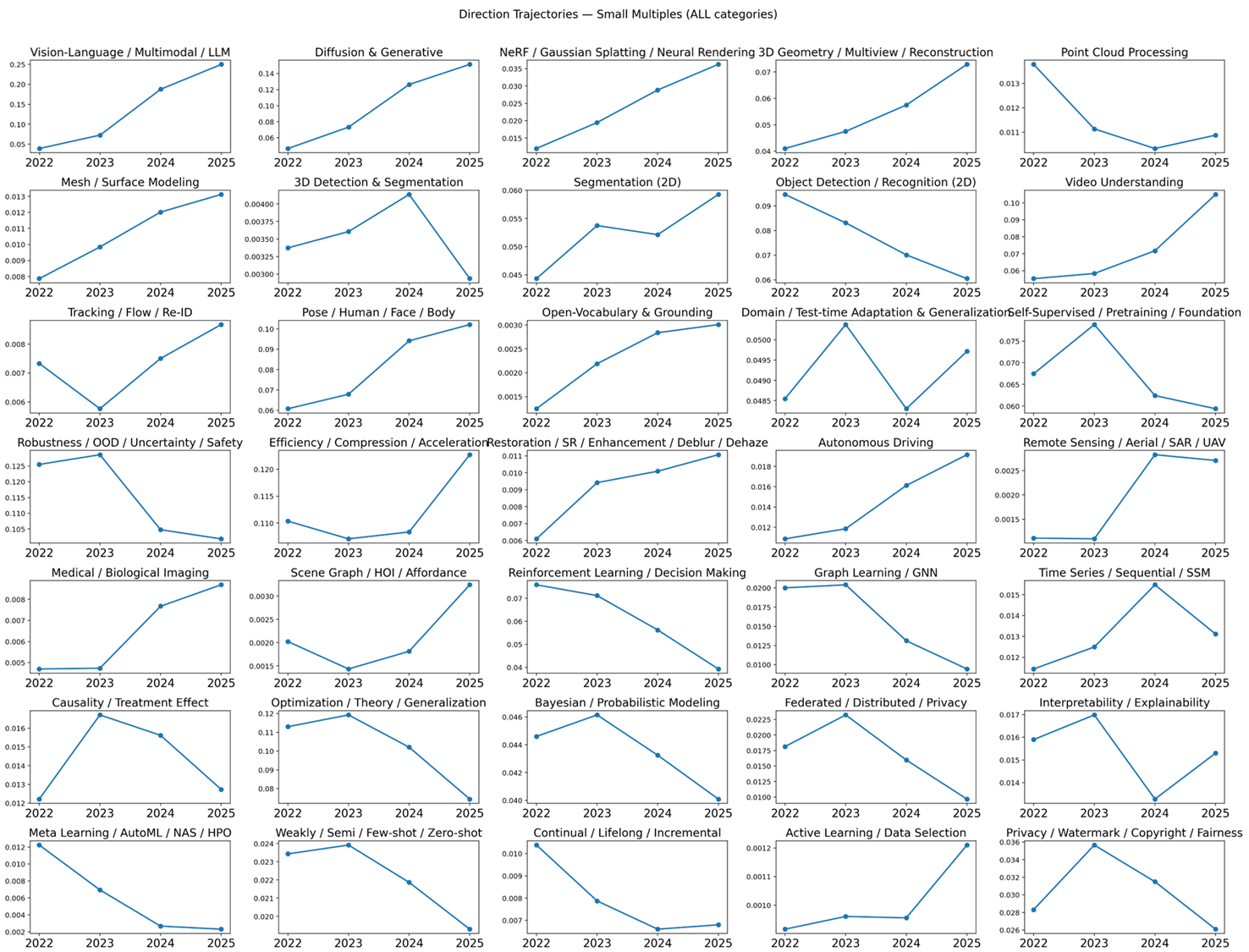}}
  \caption{\textbf{Small-multiples view of research-direction trajectories from 2022–2025}. Each panel shows one category; the x-axis is year and the y-axis is normalized topic intensity (aggregated TF–IDF score). Vision–Language/Multimodal/LLM, Diffusion \& Generative, and NeRF/Neural Rendering trend upward, while some traditional areas (e.g., 2D Object Detection, Self-Supervised/Pretraining) soften; others (e.g., GNN, Bayesian, Optimization) remain flat or decline slightly.}
  \label{fig:direction_lines_facets}
\end{figure}

Fig.\ref{fig:direction_lines_facets} presents small-multiple line charts in which each panel corresponds to a research direction. The horizontal axis shows years (2022–2025) and the vertical axis shows the fraction of all papers attributed to that direction. Because each panel has its own vertical scale, comparisons should be made within a panel across years rather than across panels by absolute height. Two global trends stand out: (i) generative and multimodal areas expand steadily and spill over into 3D, video, and editing; and (ii) several classical learning paradigms (e.g., self-supervised, meta-learning, GNNs, weak supervision) decline or plateau in relative share, while “engineering and safety” themes such as efficiency, robustness, and privacy diffuse across the field.


Generative and multimodal topics show the most pronounced rise. Diffusion and generative methods increase year over year, indicating that content generation continues to drive both methodological and application-level innovation. In parallel, vision–language and broader multimodal directions grow rapidly, reflecting the consolidation of large models and cross-modal alignment as core infrastructure. Closely related 3D content representations also climb: Gaussian splatting and neural rendering trend upward, as do 3D geometry, multiview, and reconstruction, signaling a shift from recognition alone toward high-fidelity synthesis and editable 3D assets.

Structure-aware 3D understanding strengthens as well. After an early dip, point-cloud processing rebounds slightly, while mesh and surface modeling rise steadily, suggesting interest in controllable, constraint-aware geometry. 3D detection and segmentation fluctuate: 3D detection peaks mid-period and then recedes, and 2D segmentation dips and recovers in the latest year. These oscillations likely reflect evolving benchmarks, task boundaries, and downstream deployment needs.

Temporal perception and human-centric understanding gain traction. Video understanding climbs from a low baseline, and tracking, flow, and re-identification increase gradually. Pose, face, and full-body analysis accelerate in the last two years, underscoring the move toward agent- and human-centered applications. Open-vocabulary grounding continues to rise, indicating that zero-shot recognition with language priors is becoming a standard capability.

Some “inner-loop” methodology themes cool or integrate into larger systems. Self-supervised pretraining peaks around 2023 and then declines, consistent with the field’s pivot to adapting foundation models rather than foregrounding self-supervision as a standalone contribution. Meta-learning and AutoML, weak/semi-/few-shot learning, GNNs, and causality/treatment-effect topics trend downward or remain choppy, suggesting these ideas increasingly appear as modules within broader pipelines rather than as primary focal points. Optimization and theory also edge downward, possibly due to attention shifting toward system-level integration and empirical capability building.

Engineering and reliability concerns become more visible. Efficiency, compression, and acceleration surge in the most recent year, while robustness, out-of-distribution generalization, uncertainty, and safety show steady growth. Privacy, watermarking, copyright, and fairness recede from a prior peak but maintain a notable presence, indicating that trust and governance have normalized rather than vanished. Interpretability rebounds after a dip, and federated and distributed learning stabilize after an earlier high, reflecting the ongoing importance of multi-institution collaboration and data-side constraints in real deployments.

Application-oriented areas display differentiated momentum. Medical and biological imaging rise consistently. Scene graphs, human–object interaction, and affordances strengthen recently, marking a shift from static recognition to interaction-ready understanding. Restoration, super-resolution, and enhancement track upward steadily; autonomous driving is broadly stable with a slight increase; and remote sensing maintains a modest uptick. Time-series, sequential modeling, and state-space approaches reach a high around 2024 and soften thereafter, consistent with their absorption into the foundation-model toolkit. Active learning and data selection pick up in the latest year, highlighting renewed attention to data efficiency and dataset governance.

Overall, Fig.\ref{fig:direction_lines_facets} depicts an ongoing transition toward “multimodal generative foundations plus 3D perception and editing,” while traditional paradigm-centric methods recede as independent flags and reappear as components inside larger systems. Simultaneously, scaling-aware and safety-oriented concerns grow in prominence, pushing research toward solutions that are efficient, robust, and compliant. For problem selection, the curves encourage emphasis on cross-modal, 3D, and video settings anchored in human-centric tasks, together with system designs that explicitly target efficiency, reliability, and data governance.

\begin{figure}[h]
  \centering
  \includegraphics[width=16cm]{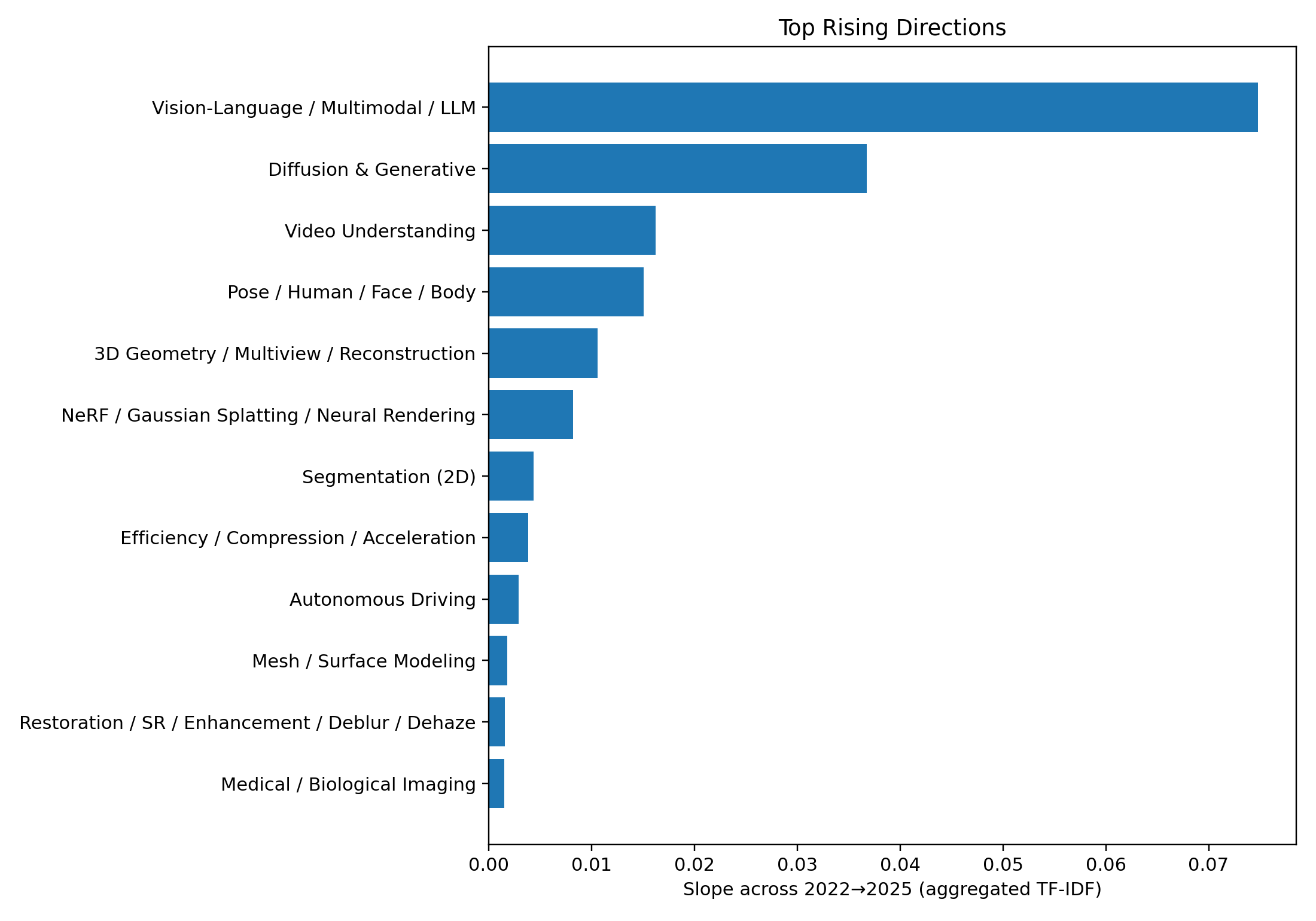}
  \caption{\textbf{Top Rising Directions across CVPR+ICLR+NeurIPS (2022--2025).}
  Bars show the slope of each direction’s aggregated TF--IDF trajectory over years.
  Larger slope = faster growth.}
  \label{fig:all-top-rising}
\end{figure}

In Fig~\ref{fig:all-top-rising}, across all venues, \emph{Vision--Language/Multimodal/LLM} exhibits the steepest increase, followed by \emph{Diffusion \& Generative}.
\emph{Video Understanding} and human-centric topics rise as well, while 3D-related areas (\emph{3D Geometry; NeRF/Gaussian Splatting}) continue to gain but at a slower rate.


\section{Vision--Language and LLMs}
\label{sec:vlm}

\textbf{Executive summary.}
Across CVPR/NeurIPS/ICLR (2023--2025), VLM abstracts pivot from
\emph{grounding/referring} toward \emph{instruction following and reasoning}.
Parameter-efficient adaptation (adapters/LoRA) and prompt-style mechanisms remain common.
Training is dominated by \emph{pretrain + finetune}, with a clear rise in \emph{instruction tuning}.
Loss design shifts away from purely contrastive objectives toward mixtures that include KL/distillation and cross-entropy/ranking.
Explicit dataset name mentions become rarer in abstracts (COCO/ImageNet steadily decline). Co-mentioned modalities tilt toward 3D and depth, while audio stabilizes and begins to recover in 2025.

\subsection{Models (named backbones \& families)}
\label{subsec:vlm-models}
\emph{ALIGN} \cite{jia2021scaling} remains the single most cited family in VLM abstracts
(\(\approx 5.8\%\) in 2024, dipping slightly to \(5.1\%\) in 2025).
\emph{LLaVA} \cite{liu2023llava} shows the fastest growth (\(0.1\%\rightarrow 1.2\%\rightarrow 2.7\%\)),
mirroring the community’s shift to instruction-following VLMs.
Classical backbones shrink in visibility—\emph{ResNet/ConvNeXt} \cite{he2016deep, liu2022convnet} and \emph{ViT} \cite{dosovitskiy2020image} roughly halve by 2025.
\emph{MoE} \cite{shazeer2017outrageously} references roughly double by 2025, indicating increasing interest in expert routing for multimodal scaling.


\begin{table}[h]
\centering
\caption{Top models referenced by VLM papers (share of VLM abstracts).}
\label{tab:vlm_models}
\begin{tabular}{lrrrrr}
\toprule
Item & 2023 & 2024 & 2025 & Trend & Slope (pp/yr) \\
\midrule
ALIGN \cite{jia2021scaling} & 4.3\% & 5.8\% & 5.1\% & −\!0.8\% & 0.65 \\
LLaVA \cite{liu2023llava} & 0.1\% & 1.2\% & 2.7\% & +2.6\% & 0.91 \\
ResNet/ConvNeXt/CNN \cite{liu2022convnet, he2016deep} & 2.9\% & 0.4\% & 0.5\% & −\!2.4\% & −0.74 \\
ViT \cite{dosovitskiy2020image} & 1.5\% & 1.2\% & 0.6\% & −\!0.9\% & −0.13 \\
MoE \cite{shazeer2017outrageously} & 0.6\% & 0.6\% & 1.3\% & +0.6\% & 0.26 \\
BLIP-2 \cite{li2023blip2} & 0.6\% & 0.2\% & 0.2\% & −\!0.4\% & 0.02 \\
BLIP \cite{li2022blip} & 0.4\% & 0.1\% & 0.1\% & −\!0.3\% & −0.00 \\
CLIP/OpenCLIP \cite{radford2021clip} & 0.1\% & 0.3\% & 0.2\% & +0.1\% & 0.06 \\
GLIP \cite{li2022grounded} & 0.4\% & 0.1\% & 0.0\% & −\!0.3\% & −0.06 \\
Swin \cite{liu2021swin} & 0.3\% & 0.1\% & 0.1\% & −\!0.2\% & −0.23 \\
Flamingo \cite{alayrac2022flamingo}  & 0.2\% & 0.1\% & 0.0\% & −\!0.2\% & −0.01 \\
GroundingDINO \cite{liu2024grounding} & 0.0\% & 0.2\% & 0.2\% & +0.2\% & 0.06 \\
\bottomrule
\end{tabular}
\end{table}

{ALIGN} \cite{jia2021scaling} (“A Large-scale ImaGe and Noisy‑text embedding”) scales contrastive dual‑encoder pretraining to {over one billion noisy image–alt‑text pairs}, deliberately avoiding heavy curation and showing that scale can compensate for label noise. A vision encoder and a text encoder are trained with a {contrastive objective} to align representations in a shared space, enabling strong \emph{zero‑shot classification} and \emph{cross‑modal retrieval} without task‑specific heads. The paper emphasizes that simple frequency filtering plus large scale suffices to surpass more complex cross‑attention models on Flickr30k/MSCOCO retrieval.

{CLIP} \cite{radford2021clip} learns image–text alignment using a {dual‑encoder} (image transformer/CNN + text transformer) trained with {InfoNCE‑style contrastive loss} on hundreds of millions of web pairs, producing a universal embedding space for \emph{zero‑shot recognition} and robust transfer. CLIP commonly serves as the {frozen vision encoder} in later LVLMs. (For scale context, BLIP reports the CLIP baseline trained on \textasciitilde\!400M pairs in its retrieval comparison table.)

{BLIP} \cite{li2022blip} (“Bootstrapping Language–Image Pretraining”) introduces a {Multimodal Mixture of Encoder–Decoder (MED)} that can act as \emph{(i)} unimodal encoders for ITC, \emph{(ii)} an image‑grounded text encoder for ITM, and \emph{(iii)} an image‑grounded {text decoder} for language modeling—jointly optimizing {ITC, ITM, and LM} to support both understanding and generation. BLIP also proposes {CapFilt} (captioning \& filtering) to \emph{bootstrap} cleaner pretraining data from noisy web alt‑text, improving retrieval, captioning, and VQA across 14M–129M images and scaling to partially include LAION.

{Flamingo} \cite{alayrac2022flamingo} is an {LVLM} that conditions a large decoder‑only LLM on visual evidence via {gated cross‑attention layers}. A lightweight {Perceiver‑Resampler} compresses image/video features into a small set of tokens fed into the language model, enabling \emph{few‑shot} multimodal learning on interleaved image–text streams. The work highlights training on a mixture of web corpora with interleaved modalities (e.g., \emph{M3W}) to yield strong few‑shot and zero‑shot VQA/retrieval.

\begin{figure*}[t]
  \centering
  \resizebox{\textwidth}{!}{
  \begin{tikzpicture}
    \draw[ultra thick, -stealth, black] (-1,0) -- (16,0);

    \foreach \x/\year in {0/2020, 1.5/2021, 3.5/2022, 7/2023, 11/2024, 12/2025} {
      \fill[black] (\x,0) circle (2pt);
      \node[below] at (\x,-0.2) {\year};
    }

    \begin{scope}[shift={(0,4.8)}]
      \draw[ultra thick, BrickRed] (-1,0)--(-0.2,0) node[right]{\small (A) Dual-encoder / Contrastive};
      \draw[ultra thick, RoyalBlue] (5,0)--(5.8,0) node[right]{\small (B) Cross-modal / Encoder--Decoder};
      \draw[ultra thick, ForestGreen] (11.5,0)--(12.3,0) node[right]{\small (C) Multimodal LLMs};
    \end{scope}

    \draw[RoyalBlue, ultra thick, -stealth] (0.5,0) -- (0.5,1)
      node[above]{\makecell[c]{UNITER \\ \cite{chen2020uniter}}};
    \draw[RoyalBlue, ultra thick, -stealth] (1,0) -- (1,-1)
      node[below]{\makecell[c]{OSCAR \\ \cite{li2020oscar}}};

    \draw[BrickRed, ultra thick, -stealth] (2,0) -- (2,2)
      node[above]{\makecell[c]{CLIP \\ \cite{radford2021learning}}};
    \draw[BrickRed, ultra thick, -stealth] (2.5,0) -- (2.5,-2)
      node[below]{\makecell[c]{ALIGN \\ \cite{jia2021scaling}}};
    \draw[RoyalBlue, ultra thick, -stealth] (3,0) -- (3,1)
      node[above]{\makecell[c]{ALBEF \\ \cite{li2021albef}}};

    \draw[RoyalBlue, ultra thick, -stealth] (4,0) -- (4,-1)
      node[below]{\makecell[c]{SimVLM \\ \cite{wang2022simvlm}}};
    \draw[RoyalBlue, ultra thick, -stealth] (4.5,0) -- (4.5,3)
      node[above]{\makecell[c]{LiT \\ \cite{zhai2022lit}}};
    \draw[RoyalBlue, ultra thick, -stealth] (5,0) -- (5,-3)
      node[below]{\makecell[c]{CoCa \\ \cite{yu2022coca}}};
    \draw[ForestGreen, ultra thick, -stealth] (5.5,0) -- (5.5,1)
      node[above]{\makecell[c]{Flamingo \\ \cite{alayrac2022flamingo}}};
    \draw[RoyalBlue, ultra thick, -stealth] (6,0) -- (6,-1)
      node[below]{\makecell[c]{BLIP \\ \cite{li2022blip}}};
    \draw[ForestGreen, ultra thick, -stealth] (6.5,0) -- (6.5,3)
      node[above]{\makecell[c]{PaLI \\ \cite{chen2022pali}}};

    \draw[ForestGreen, ultra thick, -stealth] (7.5,0) -- (7.5,-3)
      node[below]{\makecell[c]{BLIP-2 \\ (Q-Former) \\ \cite{li2023blip2}}};
    \draw[ForestGreen, ultra thick, -stealth] (8,0) -- (8,1)
      node[above]{\makecell[c]{LLaVA \\ \cite{liu2023llava}}};
    \draw[ForestGreen, ultra thick, -stealth] (8.5,0) -- (8.5,-1)
      node[below]{\makecell[c]{Instruct \\ BLIP \\ \cite{dai2023instructblip}}};
    \draw[ForestGreen, ultra thick, -stealth] (9,0) -- (9,3)
      node[above]{\makecell[c]{Kosmos-1 \\ \cite{huang2023kosmos1}}};
    \draw[ForestGreen, ultra thick, -stealth] (9.5,0) -- (9.5,-3)
      node[below]{\makecell[c]{Kosmos-2 \\ \cite{peng2023kosmos2}}};
    \draw[ForestGreen, ultra thick, -stealth] (10,0) -- (10,1)
      node[above]{\makecell[c]{MiniGPT-4 \\ \cite{zhu2023minigpt4}}};
    \draw[BrickRed, ultra thick, -stealth] (10.5,0) -- (10.5,-1)
      node[below]{\makecell[c]{OpenCLIP \\ \cite{ilharco2023openclip}}};

    \draw[ForestGreen, ultra thick, -stealth] (11.5,0) -- (11.5,3)
      node[above]{\makecell[c]{PaLI-X \\ \cite{chen2024palix}}};

    \draw[ForestGreen, ultra thick, -stealth] (12.5,0) -- (12.5, -3)
      node[below]{\makecell[c]{Qwen2.5\textendash VL \\ \cite{qwen25vl}}};
    
    \draw[ForestGreen, ultra thick, -stealth] (13,0) -- (13,1)
      node[above]{\makecell[c]{InternVL \\ 3.0 \\ \cite{internvl3}}};
    
    \draw[ForestGreen, ultra thick, -stealth] (13.5,0) -- (13.5, -1)
      node[below]{\makecell[c]{InternVL \\ 3.5 \\ \cite{internvl35}}};
    
    \draw[ForestGreen, ultra thick, -stealth] (14,0) -- (14,3)
      node[above]{\makecell[c]{LLaVA \\ OneVision\textendash 1.5 \\ \cite{llavaonevision15}}};
    
    \draw[ForestGreen, ultra thick, -stealth] (14.5,0) -- (14.5, -3)
      node[below]{\makecell[c]{Qwen3\textendash VL \\ \cite{qwen3vl}}};
    
    \draw[ForestGreen, ultra thick, -stealth] (15,0) -- (15,1)
      node[above]{\makecell[c]{Qwen3 \\ Omni \\ \cite{qwen3omni}}};
    

  \end{tikzpicture}
  }
  \caption{\textbf{Chronological overview of representative VLM / Multimodal LLM milestones (2020–2025).}
  Four color-coded categories: \textcolor{BrickRed}{(A) Dual-encoder / Contrastive},
  \textcolor{RoyalBlue}{(B) Cross-modal / Encoder--Decoder},
  \textcolor{ForestGreen}{(C) Multimodal LLMs}.
  Each arrow is vertically offset within its year and labeled above/below with the model name and citation.}
  \label{fig:vlm_timeline}
\end{figure*}
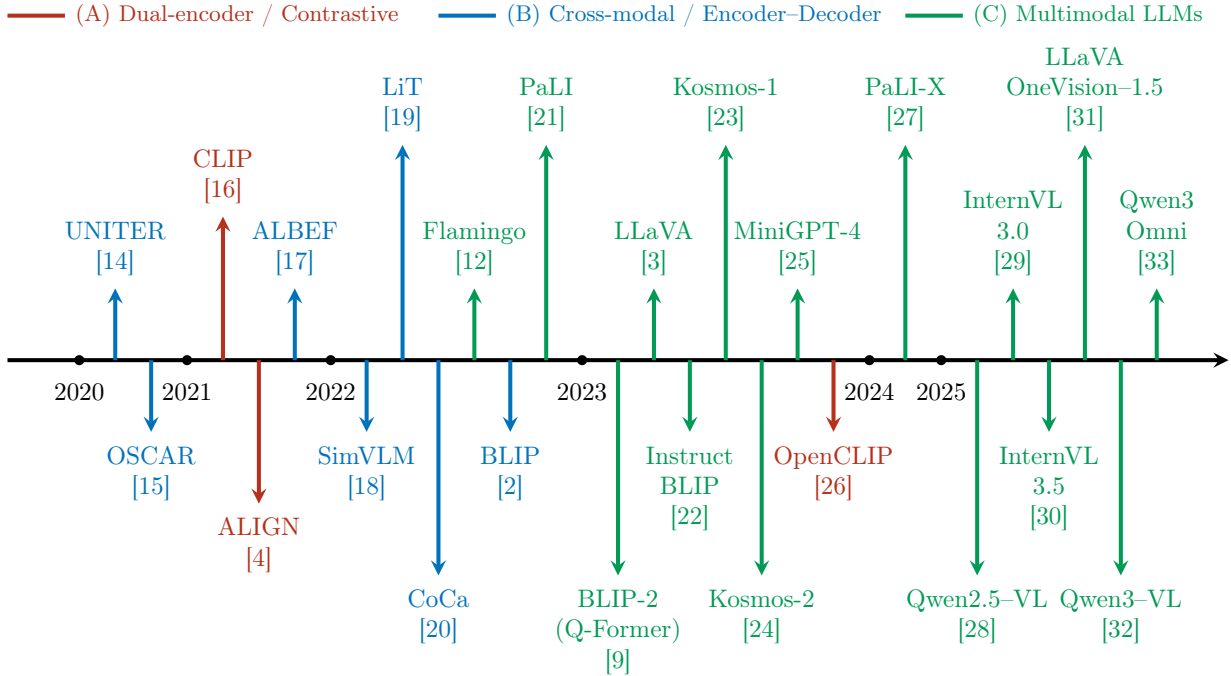

{LLaVA} \cite{liu2023llava} (“Large Language‑and‑Vision Assistant”) shows that {visual instruction tuning}—supervising an LLM with \emph{high‑quality, GPT‑4 generated multimodal conversations} paired with images—can endow the model with broad \emph{instruction following}, reasoning, and perception skills. A simple {projection bridge} maps image features from a CLIP‑like encoder into the LLM token space, after which {stage‑wise tuning} on the synthetic conversations yields strong zero‑shot generalization.

{DINO} \cite{caron2021emerging} demonstrates that {self‑distillation without labels} (student–teacher with momentum) applied to Vision Transformers yields {emergent patch‑level correspondences} and high‑quality features for downstream tasks; key ingredients are multi‑crop augmentations and centering strategies that stabilize training. DINO is often used as an initialization or backbone in vision systems. 
{DINOv2} \cite{oquab2023dinov2} scales this recipe substantially: it trains larger ViT backbones on a carefully curated, de-duplicated image corpus, strengthens the teacher–student alignment with improved regularization and prototype objectives, and targets resolution-robust, task-agnostic features. As a result, DINOv2 delivers strong zero-/few-shot transfer and competitive dense recognition performance without labels, making it a common visual encoder for VLMs and robotics pipelines.
{DINOv3} \cite{simeoni2025dinov3} further advances self-supervised ViTs by (i) scaling both data and model size with careful data preparation/optimization, (ii) introducing Gram anchoring to prevent degradation of dense feature maps under long schedules, and (iii) applying post-hoc strategies that improve flexibility across input resolution, model size, and even text-alignment. The released DINOv3 suite yields high-quality dense features and outperforms prior self-/weakly-supervised visual foundation models across a broad range of tasks.

{Grounding DINO} \cite{liu2024grounding} marries DINO‑style detectors with {open‑vocabulary pretraining}, aligning {region‑level visual features with free‑form text} so that \emph{one detector} handles both traditional object categories and arbitrary phrases (e.g., referring expressions) in a {single pipeline}. The method leverages {region–text pretraining} and {phrase grounding} signals to enable \emph{open‑set} detection.

{MoE} \cite{shazeer2017outrageously} architectures (e.g., Switch Transformers) insert a {sparse gating} mechanism that \emph{routes each token} to a small subset of specialized \emph{expert} MLPs, achieving {parameter‑efficient scaling} to hundreds of billions or trillion‑parameter regimes while keeping per‑token compute roughly constant. Such {sparse decoders} increasingly serve as the language backbone in LVLMs.

\vspace{0.5em}
Overall, across these families, two design templates dominate: {dual‑encoder contrastive} pretraining (ALIGN/CLIP) for universal embeddings, and {LLM‑centric} conditioning with cross‑attention and instruction tuning (Flamingo/LLaVA). BLIP bridges understanding–generation with its MED and \emph{data bootstrapping}, while DINO and Grounding‑DINO provide strong {vision backbones} and {open‑vocabulary grounding} that LVLMs can exploit.

\begin{table}[h]
\centering
\caption{\textbf{Named model families and design choices.} We summarize core ideas, fusion styles, typical backbones, objectives, and the nature/scale of pretraining data when explicitly reported in the cited papers. Abbreviations: ITC/ITM = image–text contrast / matching; LM = language modeling; GCA = gated cross-attention.}
\label{tab:model_family_compare}

\resizebox{\textwidth}{!}{%
\begin{minipage}{\textwidth}
\small
\begin{tabularx}{\textwidth}{
  L{1.8cm}  
  Y          
  L{2.0cm}   
  Y          
  L{2.8cm}   
  L{2.8cm}   
}
\toprule
\textbf{Model} & \textbf{Core Idea} & \textbf{Fusion} & \textbf{Vision \& Text Backbones} & \textbf{Objective(s)} & \textbf{Pretraining Data}\\
\midrule
ALIGN \cite{jia2021scaling}& Dual-encoder at web scale & None (late) & CNN/Transformer encoders & Contrastive alignment & $\sim$1B+ noisy alt-text pairs \\
\addlinespace[0.25em]
CLIP \cite{radford2021clip}& Dual-encoder, universal embeddings & None (late) & ViT/ResNet + text Transformer & Contrastive (InfoNCE) & $\sim$400M web pairs \\
\addlinespace[0.25em]
BLIP \cite{li2022blip}& MED unifies understanding \& generation + CapFilt & Cross-attn in encoder/decoder & ViT + BERT-style text Transformer & ITC + ITM + LM & 14M–129M images (+ partial LAION) \\
\addlinespace[0.25em]
Flamingo \cite{alayrac2022flamingo}& LVLM with GCA layers and Perceiver Resampler & Gated cross-attention & CNN/ViT features $\rightarrow$ LLM decoder & Autoregressive LM (vision-conditioned) & Interleaved image–text corpora (e.g., M3W) \\
\addlinespace[0.25em]
LLaVA \cite{liu2023llava}& Visual instruction tuning with GPT-4 conversations & Projector $\rightarrow$ LLM tokens & CLIP-like image encoder + decoder LLM & SFT / LM on multimodal dialogs & High-quality synthetic conversations \\
\addlinespace[0.25em]
DINO \cite{caron2021emerging}& Self-distillation ViT without labels & N/A & Vision Transformer (student/teacher) & Self-distill with momentum teacher & Unlabeled images (multi-crop) \\
\addlinespace[0.25em]
Grounding DINO \cite{liu2024grounding}& Open-vocabulary detection by region–text pretraining & Cross-modal at region level & Detector + text encoder (phrase grounding) & Detection + grounding losses & Region–phrase corpora (web \& annotations) \\
\addlinespace[0.25em]
MoE \cite{shazeer2017outrageously}& Token routing to expert MLPs (sparse) & Inside decoder (sparse) & Decoder-only Transformer with experts & Autoregressive LM, sparse gating & Large text corpora (scalable) \\
\bottomrule
\end{tabularx}
\end{minipage}%
}
\end{table}

\vspace{0.5em}
In Table~\ref{tab:model_family_compare},
\textbf{Dual‑Encoder} families (ALIGN/CLIP) emphasize \textbf{dataset scale} and a \textbf{contrastive} recipe for broad transfer. \textbf{LVLM} families (Flamingo/LLaVA) inject vision into a powerful language decoder via \textbf{cross‑attention} or a simple \textbf{projector}, and then rely on \textbf{instruction‑like supervision} to elicit reasoning/grounding. \textbf{BLIP} occupies the middle ground by unifying \textbf{ITC/ITM/LM} and \textbf{bootstrapping} cleaner captions. \textbf{DINO} and \textbf{Grounding DINO} secure the vision side—self‑supervised features and open‑vocabulary localization—that downstream LVLMs frequently build upon. Finally, \textbf{MoE} provides a path to scale \textbf{language backbones} with sparse compute, increasingly common in recent LVLMs.


\subsection{Fusion / Architectural Integration}
\label{subsec{vlm-fusion}}

Table~\ref{tab:vlm_fusion} summarizes how VLM papers integrate vision and language components. We observe three clear movements.

Lightweight promptization rises. Prompt/Prefix tuning is the most frequently referenced mechanism and continues to trend upward (13.0,$\rightarrow$,16.4,$\rightarrow$,14.3\%, Trend $+1.3$,pp; Slope $+3.43$,pp/yr), reflecting a preference for parameter‐efficient adaptation on strong frozen backbones~\cite{lester2021prompt,li2021prefix}. Adapter/LoRA usage also grows steadily (1.3,$\rightarrow$,4.0,$\rightarrow$,4.1\%, Trend $+2.8$,pp), consistent with the widespread adoption of low-rank updates for multimodal fine-tuning~\cite{hu2022lora}.

Bridging modules remain active while “heavy” fusion softens. Cross-/co-attention stays stable to slightly positive (Trend $+0.5$,pp), often appearing in systems that inject visual features into frozen LMs (e.g., Flamingo-style cross-attention layers)~\cite{alayrac2022flamingo}. Projector/MLP heads increase (Trend $+0.6$,pp), consistent with simple alignment layers that map modality embeddings into a shared space (as in CLIP-style dual encoders with learned projections)~\cite{radford2021clip}. The Q-Former bridge (BLIP-2) stays non-negligible and flat (Trend $+0.0$,pp), suggesting continued but specialized use where learned query tokens are advantageous~\cite{li2023blip2}.

Architecture choices rebalance. Mixture-of-Experts/gating is mildly up (Trend $+0.8$,pp), indicating exploratory interest in sparse capacity~\cite{shazeer2017moe}. By contrast, encoder–decoder patterns show a small net decline (Trend $-1.3$,pp), despite remaining important for generation-centric systems (e.g., OFA, PaLI)~\cite{wang2022ofa,chen2022pali}. Dual-encoder/two-tower usage edges downward (Trend $-0.1$,pp), likely reflecting a shift from pure retrieval/contrastive training toward instruction-following and generative pipelines where cross-attention or bridging modules provide tighter fusion.

\begin{table}[h]
\centering
\caption{\textbf{Fusion/architectural integration mechanisms within VLM papers}. “Trend” is the three-year net change (2025 minus 2023, in percentage points). “Slope” is the least-squares linear slope across 2023–2025.}
\label{tab:vlm_fusion}
\begin{tabular}{lrrrrr}
\toprule
Item & 2023 & 2024 & 2025 & Trend & Slope (pp/yr) \\
\midrule
Prompt/Prefix Tuning & 13.0\% & 16.4\% & 14.3\% & +1.3\% & 3.43 \\
Adapter/LoRA & 1.3\% & 4.0\% & 4.1\% & +2.8\% & 1.26 \\
Cross-/Co-attention & 1.7\% & 2.2\% & 2.2\% & +0.5\% & 0.46 \\
Projector/MLP Head & 0.9\% & 1.2\% & 1.5\% & +0.6\% & 0.21 \\
MoE/Gating & 0.6\% & 0.6\% & 1.4\% & +0.8\% & 0.24 \\
Encoder-Decoder & 1.6\% & 0.7\% & 0.3\% & −\!1.3\% & −0.39 \\
Dual-encoder/Two-tower & 0.3\% & 0.2\% & 0.1\% & −\!0.1\% & −0.04 \\
Q-Former Bridge & 0.0\% & 0.1\% & 0.0\% & +0.0\% & 0.03 \\
\bottomrule
\end{tabular}
\end{table}

Overall, the field is converging on parameter-efficient tuning + light bridging as default design knobs (for cost, stability, and reusability), while reserving cross-attention/Q-Former for cases that require stronger, token-level conditioning. Sparse MoE appears as an emerging capacity-scaling option, whereas monolithic encoder–decoder and pure dual-encoder designs, though still influential, grow more selectively applied.


\subsection{Tasks}
\label{subsec:vlm-tasks}

Table~\ref{tab:vlm_tasks_new} profiles how task emphases have evolved in recent VLM work (shares over 2023–2025). We see a decisive pivot toward instruction-following and multi-step reasoning, alongside a broad cooling of classic grounding/captioning style tasks.

\begin{table}[h]
\centering
\caption{Task strata within VLM papers (share).}
\label{tab:vlm_tasks_new}
\begin{tabular}{lrrrrr}
\toprule
Item & 2023 & 2024 & 2025 & Trend & Slope (pp/yr) \\
\midrule
Reasoning/Instruction & 13.5\% & 22.3\% & 25.0\% & +11.5\% & 5.71 \\
Grounding/Referring & 25.9\% & 14.5\% & 12.9\% & −\!13.0\% & −8.36 \\
Retrieval & 8.5\% & 6.8\% & 8.3\% & −\!0.2\% & 0.53 \\
Captioning & 6.2\% & 4.9\% & 4.4\% & −\!1.9\% & −0.53 \\
VQA & 2.4\% & 2.0\% & 1.9\% & −\!0.5\% & −0.05 \\
Video QA/Captioning & 1.3\% & 1.0\% & 1.2\% & −\!0.1\% & 0.02 \\
OCR/Text Recognition & 0.9\% & 1.0\% & 1.0\% & +0.2\% & −0.44 \\
Open-Vocabulary Det./Seg. & 1.4\% & 0.8\% & 0.4\% & −\!1.0\% & −0.12 \\
\bottomrule
\end{tabular}
\end{table}

Reasoning / Instruction: The fastest-growing stratum rises from 13.5\% to 25.0\% (Trend $+11.5$,pp; Slope $+5.71$,pp/yr), driven by instruction-tuned LMMs that attach visual adapters to frozen (or lightly fine-tuned) LMs and are optimized with conversation-style data (e.g., LLaVA, InstructBLIP, MiniGPT-4)~\cite{liu2023llava,dai2023instructblip,zhu2023minigpt4}. These systems prioritize chain-of-thought localization, tool-use hooks, and preference optimization, naturally expanding the proportion of “reasoning” papers.

Grounding / Referring: Once dominant, this stratum declines by 13.0,pp overall, reflecting a shift from RefCOCO/phrase-grounding formulations toward instruction-following stacks where grounding appears as a sub-capability rather than the end task. Methodologically, open-vocabulary detectors (e.g., GLIP, GroundingDINO) remain key components but are now more often embedded as plug-ins~\cite{li2022grounded,liu2024grounding}.

Retrieval: The share is broadly stable (8.5,$\rightarrow$,8.3\%), consistent with the continued role of dual-encoders for retrieval and RAG back-ends (CLIP-style alignment; BLIP/OFA hybrids for generative read-out)~\cite{radford2021clip,li2022blip,wang2022ofa}.

Captioning and VQA: Generic image captioning declines modestly ($-1.9$,pp), likely absorbed by instruction-chat settings where captioning becomes an intermediate capability~\cite{li2022blip}. VQA dips slightly ($-0.5$,pp), as community benchmarks (e.g., VQA v2, GQA) give way to broader multimodal instruction suites; nonetheless VQA remains a standard probe for perception-reasoning coupling~\cite{goyal2017vqa,hudson2019gqa}.

Video QA / Captioning: The slice is nearly flat, indicating steady but not explosive growth. While datasets such as MSRVTT-QA and NExT-QA continue to anchor evaluation, compute/data cost keeps video tasks a smaller share relative to image-centric reasoning~\cite{xu2017msrvttqa,xiao2021nextqa}.

OCR / Text recognition: The share is small but persistent (roughly 1\%), reflecting increased demand for document-centric VLMs (e.g., TrOCR/Donut lines) and the prevalence of text-rich scenes within instruction datasets~\cite{li2021trocr,kim2022donut}.

Open-vocabulary detection / segmentation: A gradual decline (−1.0,pp) likely reflects consolidation: many recent LMMs \emph{use} open-vocab perception as a front-end (OWL-ViT/ViLD/SAM/OpenSeg families) rather than publishing it as a standalone task~\cite{minderer2022owlvit,gu2021vild,kirillov2023segment,ghiasi2022open}.

Overall, the center of gravity is moving from task-specific supervision (grounding, captioning) toward generalist, instruction-tuned reasoning with retrieval and open-vocab perception as composable sub-systems. This aligns with the design trend noted in Sec.~\ref{subsec{vlm-fusion}}: lightweight adaptation of powerful frozen LMs plus modular visual front-ends.


\subsection{Training Paradigms}
\label{subsec:vlm-train}

Table~\ref{tab:vlm_train_new} tracks the training regimes emphasized in recent VLM papers (2023–2025). We observe a decisive shift toward
\emph{fine-tuning strong pretrained components with parameter-efficient knobs and instruction data}, while earlier web-scale weak/ self-supervision plays a relatively smaller role in new papers.

\begin{table}[h]
\centering
\caption{Training paradigms appearing in VLM papers.}
\label{tab:vlm_train_new}
\begin{tabular}{lrrrrr}
\toprule
Item & 2023 & 2024 & 2025 & Trend & Slope (pp/yr) \\
\midrule
Pretrain + Finetune & 11.6\% & 16.9\% & 16.8\% & +5.2\% & 3.83 \\
Prompt/Prefix & 13.0\% & 16.4\% & 14.3\% & +1.3\% & 3.43 \\
Self-/Weak-/Semi- sup. & 9.6\% & 2.8\% & 3.5\% & −\!6.1\% & −2.65 \\
Distillation & 4.2\% & 4.8\% & 4.0\% & −\!0.4\% & 0.56 \\
Instruction Tuning & 1.1\% & 4.2\% & 5.0\% & +3.9\% & 1.75 \\
LoRA/Adapters & 1.3\% & 4.0\% & 4.1\% & +2.8\% & 1.26 \\
Multi-task/Curriculum & 2.6\% & 1.6\% & 1.9\% & −\!0.8\% & −0.08 \\
\bottomrule
\end{tabular}
\end{table}

Pretrain + Finetune:
This remains the anchor recipe and grows by ${+}5.2$\,pp overall. Typical patterns couple a frozen or lightly finetuned LM with an image encoder that was pretrained via contrastive/web supervision (e.g., CLIP/ALIGN), then finetune end-to-end or at the bridge for downstream tasks~\cite{radford2021clip,jia2021scaling,li2022blip}.

Prompt / Prefix:
Promptization rises (${+}1.3$\,pp), reflecting the appeal of \emph{tuning a small prompt vector} instead of full weights for new domains or tasks~\cite{lester2021prompt,li2021prefix}. In multimodal stacks this often means learning visual prompts or cross-modal prompts while keeping the LM mostly frozen.

Instruction Tuning:
A substantial increase (${+}3.9$\,pp) mirrors the community pivot to conversational VLMs: systems such as LLaVA and InstructBLIP attach vision adapters and then perform supervised instruction tuning on (often synthetic) multimodal dialogues~\cite{liu2023llava,dai2023instructblip}. This training style turns captioning/grounding capabilities into general instruction following.

LoRA / Adapters:
Parameter-efficient updates expand (${+}2.8$\,pp), driven by their low compute/VRAM cost and stability when merging or stacking skills across domains~\cite{hu2022lora,houlsby2019adapter}. In practice, many VLMs combine LoRA with instruction tuning.

Self- / Weak- / Semi-supervision:
The share decreases (${−}6.1$\,pp). While self/weak supervision (e.g., SimCLR, MoCo; large-scale noisy web pairs as in ALIGN/CLIP) remains crucial for \emph{pretraining} encoders, new papers increasingly take such backbones as given and focus on finetuning/instruction stages instead~\cite{chen2020simclr,he2020moco,radford2021clip,jia2021scaling,xie2020noisystudent}.

Distillation:
Usage is relatively steady (net ${−}0.4$\,pp; slight positive slope), often appearing to compress vision encoders, align modality bridges, or transfer reasoning supervision from stronger teachers into smaller student LMMs~\cite{hinton2015distill,li2022blip}. 

Multi-task / Curriculum:
A small decline (${−}0.8$\,pp) likely reflects consolidation: unified instruction tuning increasingly subsumes the benefits of multi-task/curriculum training observed in earlier generalist frameworks such as OFA/PaLI and instruction-style curricula (FLAN)~\cite{wang2022ofa,chen2022pali,wei2022flan}.

Overall, the center of gravity moves from building encoders with massive weak supervision to \emph{adapting} those encoders (and frozen LMs) with instruction data and parameter-efficient updates. This reduces cost, speeds iteration, and aligns with the modular fusion trends in Sec.~\ref{subsec{vlm-fusion}}.


\subsection{Loss Families}
\label{subsec:vlm-loss}

Table~\ref{tab:vlm_losses_new} summarizes the loss objectives most commonly reported in VLM papers from 2023–2025. We observe a
\emph{rebalancing from pretraining-style contrastive objectives toward alignment, supervision, and compression losses} that better fit
instruction-tuned, generation-oriented pipelines.

\begin{table}[h]
\centering
\caption{Loss families in VLM papers.}
\label{tab:vlm_losses_new}
\begin{tabular}{lrrrrr}
\toprule
Item & 2023 & 2024 & 2025 & Trend & Slope (pp/yr) \\
\midrule
Contrastive/InfoNCE & 10.8\% & 5.6\% & 5.1\% & −\!5.7\% & −2.07 \\
KL/Distillation & 5.6\% & 6.6\% & 5.8\% & +0.3\% & 0.78 \\
Triplet/Ranking & 1.0\% & 0.7\% & 0.5\% & −\!0.5\% & −0.00 \\
Cross-Entropy/Focal & 0.8\% & 0.3\% & 0.6\% & −\!0.1\% & −0.19 \\
MSE/L1/L2 & 0.3\% & 0.4\% & 0.3\% & −\!0.0\% & −0.10 \\
Dice/IoU & 0.4\% & 0.2\% & 0.1\% & −\!0.3\% & −0.06 \\
Chamfer/EMD & 0.1\% & 0.2\% & 0.0\% & −\!0.1\% & −0.10 \\
\bottomrule
\end{tabular}
\end{table}

Contrastive / InfoNCE:
The share of contrastive learning drops markedly ($-5.7$\,pp; slope $-2.07$\,pp/yr). This mirrors a shift in new works away from building
image--text encoders from scratch (as in CLIP/ALIGN) toward \emph{adapting} such pretrained encoders and coupling them with large LMs.
Contrastive learning remains foundational for representation learning and retrieval~\cite{oord2018cpc,chen2020simclr,radford2021clip},
but appears less frequently as the \emph{primary} objective in papers that focus on multimodal instruction following.

KL / Distillation:
A small net increase ($+0.3$\,pp) reflects steady use of Kullback–Leibler–based distillation to compress vision backbones, align vision--language
bridges, or transfer reasoning signals from stronger LMM teachers to lighter students~\cite{hinton2015distill,li2022blip}. Distillation is
often combined with supervised instruction losses to stabilize training.

Triplet / Ranking:
This family remains low and stable. Triplet and margin-based ranking continue to appear in retrieval and grounding submodules,
but are less central when the end task is dialogue-style generation~\cite{schroff2015facenet}.

Cross-Entropy / Focal:
Classification-style objectives are modestly represented. They are widely used for detector heads or token-level supervision in
bridging modules; focal loss is occasionally adopted for long-tail/open-vocabulary settings~\cite{lin2017focal}.

MSE / L1 / L2:
Regression losses persist at a low but steady rate, mainly for projector heads, alignment regressors, or value heads in reward/prefence
optimization stages. Their prevalence is limited by the dominance of token-level cross-entropy in instruction tuning.

Dice / IoU:
These segmentation-oriented objectives shrink slightly as standalone open-vocabulary perception becomes a plug-in capability, rather than
a primary contribution in VLM papers~\cite{milletari2016vnet,rahman2016iou}.

Chamfer / EMD:
Point-set distances (Chamfer, Earth Mover's) are niche in mainstream VLM work and mostly appear in papers that also target 3D grounding or
generative geometry, where set/shape supervision is required~\cite{fan2017pointset}.

Overall, compared to 2023, losses in 2025 VLM papers reflect a community focus on \emph{instruction-tuned generation with modular perception}:
contrastive losses are still crucial for encoders and retrieval, but many new contributions emphasize KL-based distillation and CE-style
supervision layered on top of pretrained backbones and LMs.


\subsection{Datasets}
\label{subsec:vlm-data}

Table~\ref{tab:vlm_datasets_new} lists curated datasets that authors explicitly mention in VLM abstracts. We caution that abstracts under-report training data, especially for recent generalist LMMs that rely on mixtures of private or filtered corpora. Even with that caveat, three patterns emerge. First, legacy caption benchmarks such as MS-COCO~\cite{lin2014coco,chen2015cococap} and ImageNet~\cite{deng2009imagenet} decline steadily in mentions (COCO: $-3.0$\,pp; ImageNet: $-1.5$\,pp). They remain common for sanity checks and zero-shot probes, but fewer new papers position them as primary contributions. Second, the share of open-web sources is small but persistent: LAION~\cite{schuhmann2021laion400m,schuhmann2022laion5b} stays roughly flat overall, consistent with a trend where authors reference large web datasets generically while focusing the abstract on downstream instruction tuning. Third, task-specific datasets shrink or stabilize, reflecting a shift toward instruction-chat evaluations where many traditional tasks appear as sub-skills rather than endpoints. RefCOCO/RefCOCO+/RefCOCOg for referring expressions~\cite{yu2016refcoco,mao2016refcocog}, Flickr30k for captioning~\cite{young2014flickr30k}, CC3M/CC12M for web-scale caption pretraining~\cite{sharma2018cc3m,changpinyo2021cc12m}, VQA-v2 and OK-VQA for QA and knowledge-augmented QA~\cite{goyal2017vqa,marino2019okvqa}, WebVid/MSR-VTT/MSVD for video-text pairs~\cite{bain2021webvid,xu2016msrvtt,chen2011msvd}, YouCook2/HowTo100M for instructional video grounding~\cite{zhou2018youcook2,miech2019howto100m}, and Visual Genome for dense region-language grounding~\cite{krishnavisualgenome2017} all show flat to negative trends in abstract mentions.

\begin{table}[h]
\centering
\caption{Curated datasets explicitly named in VLM abstracts (note under-reporting).}
\label{tab:vlm_datasets_new}
\begin{tabular}{lrrrrr}
\toprule
Item & 2023 & 2024 & 2025 & Trend & Slope (pp/yr) \\
\midrule
MS-COCO & 4.9\% & 2.1\% & 1.0\% & −\!3.0\% & −1.50 \\
ImageNet & 3.1\% & 2.4\% & 1.6\% & −\!1.5\% & −0.76 \\
LAION & 0.6\% & 0.8\% & 0.2\% & −\!0.5\% & 0.03 \\
RefCOCO/g/+ & 0.6\% & 0.6\% & 0.3\% & −\!0.3\% & −0.12 \\
Flickr30k & 0.8\% & 0.3\% & 0.2\% & −\!0.7\% & −0.28 \\
CC3M/CC12M & 0.4\% & 0.3\% & 0.3\% & −\!0.1\% & −0.19 \\
VQA-v2/OK-VQA & 0.4\% & 0.2\% & 0.3\% & −\!0.2\% & −0.09 \\
WebVid/MSRVTT/MSVD & 0.3\% & 0.3\% & 0.1\% & −\!0.2\% & −0.04 \\
YouCook2/HowTo & 0.2\% & 0.2\% & 0.1\% & −\!0.1\% & −0.16 \\
Visual Genome & 0.3\% & 0.1\% & 0.0\% & −\!0.2\% & −0.05 \\
COCO Captions & 0.1\% & 0.0\% & 0.0\% & −\!0.0\% & −0.03 \\
\bottomrule
\end{tabular}
\end{table}

Overall, abstracts increasingly de-emphasize named, single-dataset training and emphasize model behavior on multi-task, instruction-style suites. This aligns with the broader movement in Sections~\ref{subsec{vlm-fusion}} and~\ref{subsec:vlm-train} toward reusing strong pretrained encoders, lightweight adaptation, and instruction tuning, with classic datasets retained primarily for probing, ablations, and comparability.


\subsection{Modalities (co-mentioned with VLM)}
\label{subsec:vlm-modality}

Table~\ref{tab:vlm_modal_new} tracks which \emph{additional} modalities are co-mentioned alongside vision–language modeling in 2023–2025 abstracts. Three observations emerge. First, 3D/point–cloud signals tick upward overall (Trend $+0.7$\,pp), echoing growing interest in grounding LMMs to embodied/3D settings (robotics, digital twins, scene understanding). Much of this activity aligns features across modalities rather than building fully new architectures, e.g., shared embedding spaces that can accept RGB, depth, audio, IMU, or 3D point inputs as in ImageBind’s multi-sense alignment~\cite{girdhar2023imagebind}. Second, classic image–text mentions decline (−3.5\,pp), consistent with a shift from single-pair supervision toward broader instruction-style training where images appear interleaved with other cues (video frames, depth maps, layout, etc.). Depth/RGB-D is roughly steady to slightly positive (−0.2\,pp net but a small positive slope), reflecting practical pipelines that fuse RGB with monocular or sensor depth for perception and 3D reasoning, again typically through lightweight adapters or shared-token bridges rather than bespoke RGB-D encoders. Third, audio/speech and video–text show modest declines (−1.4\,pp and −1.1\,pp, respectively). These modalities remain active, but many new papers focus on generalist instruction-following while citing prior audio/video-capable foundations instead of proposing stand-alone audio- or video-specific objectives; e.g., Flamingo-style cross-attention handles interleaved frames and extends naturally to audio features when available~\cite{alayrac2022flamingo}.

\begin{table}[h]
\centering
\caption{Additional modalities co-mentioned in VLM papers.}
\label{tab:vlm_modal_new}
\begin{tabular}{lrrrrr}
\toprule
Item & 2023 & 2024 & 2025 & Trend & Slope (pp/yr) \\
\midrule
3D/Point Cloud & 10.7\% & 11.7\% & 11.4\% & +0.7\% & 0.71 \\
Image-Text & 8.6\% & 6.2\% & 5.1\% & −\!3.5\% & −0.42 \\
Depth/RGB-D & 4.7\% & 4.1\% & 4.6\% & −\!0.2\% & 0.22 \\
Audio/Speech & 5.4\% & 3.5\% & 4.0\% & −\!1.4\% & −0.99 \\
Video-Text & 2.7\% & 1.5\% & 1.6\% & −\!1.1\% & −0.31 \\
\bottomrule
\end{tabular}
\end{table}

Overall, the co-mention trends suggest a pragmatic strategy: reuse strong image–text roots and \emph{attach} additional modalities via alignment or prompting, reserving heavy bespoke modeling only where task benefits are clear (e.g., robotics 3D grounding or long-horizon video understanding). This is consistent with the broader design and training shifts reported in Sections~\ref{subsec{vlm-fusion}} and~\ref{subsec:vlm-train}.

\paragraph{Takeaways for readers about VLMs.}
(1) If your problem can be framed as \emph{instruction} or \emph{reasoning}, emphasize it in the title/abstract to align with the strongest tailwinds.
(2) Parameter-efficient \emph{adapters/LoRA} and prompt-style interfaces have become lingua franca—treat them as strong baselines or ablations.
(3) When contrastive pretraining is not the core contribution, keep it minimal and focus on downstream instruction/finetuning strategy and broad evaluation.
(4) Explicit dataset name-dropping is rarely needed unless the dataset itself is a contribution; for video/3D, clarifying scale and diversity is more informative.
(5) If you use 3D/depth/audio, state it early; cross-modal grounding is increasingly valued.

\section{Cross-venue Comparison and Practical Advice}
\textbf{Cross‑venue comparison.} CVPR retains the strongest 3D emphasis (e.g., 23.1\% of 2025 CVPR abstracts mention 3D geometry vs. 7.8\% at ICLR), while ICLR has the largest 2025 VLM share (40.7\%, comparable to CVPR’s 39.5\%). CVPR sees the highest diffusion share in 2025 (25.7\%) among the three. NeurIPS—available up to 2024 in our data—shows an early VLM ramp (30.5\% in 2024) with diffusion at ~11.6\%.
\newline
\newline
\textbf{Actionable advice for readers.} 
(1) \emph{Connect perception with VLMs.} Successful 2025 papers increasingly formulate classic vision problems (detection, segmentation, tracking) as instruction‑following, grounding, or tool‑using tasks on top of pretrained multimodal backbones. Designing lean projectors/adapters and strong data recipes (instruction or preference alignment) is impactful. 
(2) \emph{Make generative tools usable.} If diffusion is part of the pipeline, emphasize controllability, speed/distillation, and reliability; benchmarks with precise quality/latency trade‑offs resonate with current trends. 
(3) \emph{Think long‑context and video.} Methods that scale to minute‑ or hour‑long sequences while preserving reasoning ability and memory efficiency are rising. 
(4) \emph{Be explicit about efficiency and safety.} Lightweight inference, sparsity, cache‑aware design, and safety/robustness concerns are common acceptance signals across venues.

\section{Limitations}
Our pipeline is lexicon‑driven on abstracts only. Some fields (e.g., datasets, losses) are systematically under‑reported in abstracts; therefore the absolute numbers are conservative. Papers may be multi‑label; percentages are fractions of all papers per year, not summing to 100\%. Nonetheless, the main trends (VLM ascent, diffusion pragmatization, steady 3D, rising video) are robust across venues and years.

\section{Conclusion}
We provide a transparent, replicable measurement of what the community has worked on in 2023–2025. 
By 2025, multimodal VLMs have become the organizing center for a large portion of accepted papers; diffusion matured into controllable, accelerated modules; and 3D and video remain vibrant with more cross‑modal formulations. 
We release the full lexicon and code (in the conversation history) to encourage reproducibility and extension to other venues/years.

\bibliographystyle{unsrt}
\bibliography{references}

\end{document}